\documentclass{article}

\usepackage[final]{neurips_2019}

\usepackage[utf8]{inputenc} %
\usepackage[T1]{fontenc}    %
\usepackage{hyperref}       %
\usepackage{url}            %
\usepackage{booktabs}       %
\usepackage{amsfonts}       %
\usepackage{nicefrac}       %
\usepackage{microtype}      %

\usepackage{multirow}
\usepackage{floatflt}
\usepackage{graphicx}
\usepackage{subcaption}
\usepackage{amsmath}
\usepackage{enumitem}

\usepackage[title]{appendix}

\usepackage[table,xcdraw]{xcolor}
\definecolor{mydarkblue}{rgb}{0,0.08,0.45}

\hypersetup{ %
    pdftitle={Cost-Sensitive Training for Autoregressive Models},
    pdfauthor={Irina Saparina, Anton Osokin},
    pdfsubject={Cost-Sensitive Training for Autoregressive Models},
    pdfkeywords={machine learning, autoregressive models, machine translation},
    pdfborder=0 0 0,
    pdfpagemode=UseNone,
    colorlinks=true,
    linkcolor=mydarkblue,
    citecolor=mydarkblue,
    filecolor=mydarkblue,
    urlcolor=mydarkblue,
    pdfview=FitH
}

\setdescription{leftmargin=0em, labelindent=0em}

\graphicspath{{./img/}}

\title{Cost-Sensitive Training for Autoregressive Models}

\author{%
  Irina Saparina \qquad \qquad Anton Osokin\\
  National Research University Higher School of Economics, Samsung-HSE lab\\ Moscow, Russia\\
}

\begin{document}

\maketitle

\begin{abstract}

Training autoregressive models to better predict under the test metric, instead of maximizing the likelihood, has been reported to be beneficial in several use cases but brings additional complications, which prevent wider adoption.
In this paper, we follow the learning-to-search approach~\citep{Daume2009b,searnn2018leblond} and investigate its several components.
First, we propose a way to construct a reference policy based on an alignment between the model output and ground truth.
Our reference policy is optimal when applied to the Kendall-tau distance between permutations (appear in the task of word ordering) and helps when working with the METEOR score for machine translation. 
Second, we observe that the learning-to-search approach benefits from  choosing the costs related to the test metrics. 
Finally, we study the effect of different learning objectives and find that the standard KL loss only learns several high-probability tokens and can be replaced with ranking objectives that target these tokens explicitly.

\end{abstract}

\section{Introduction}
Autoregressive models are a popular choice for many applications, including machine translation, image captioning and code generation. These models output predictions one by one, and each prediction depends on the previous ones (by prediction we mean choosing a token from the set of available tokens).
Modern autoregressive models are usually trained with the maximum likelihood estimation  (MLE) approach, which has at least two disturbing properties: exposure bias \citep{Ranzato2016b} and loss-evaluation mismatch \citep{Wiseman2016b}.
\begin{description}
   \item[Exposure bias:] 
 training with the MLE objective requires the model to take  ground-truth sequences as inputs but, to generate the output sequence step-by-step at the testing stage, the model takes its output from the previous steps as the input.
This discrepancy implies that the model never sees its own errors at the training stage, thus never explicitly learns to correct them.
   \item[Loss-evaluation mismatch:] the MLE approach attempts to learn a full probabilistic model of the ground truth, which requires enormous  diverse datasets and heuristic prediction algorithms \citep{Stahlberg2019}. 
Learning directly to predict under the target cost function might require  less data and be better compatible with available prediction algorithms. 
\end{description}

These properties have motivated \citet{Bahdanau2016, edunov-etal-2018-classical, zhang-etal-2019-bridging} to work on new training methods.
In this paper, we follow the learning-to-search (L2S) line of work \citep{Daume2009b, Chang2015, searnn2018leblond}.
This approach performs a reduction of sequence prediction to the cost-sensitive multi-class classification, where the classes  correspond  to  the tokens available at each individual prediction. 
The costs of the classes come from the metrics between the ground-truth sequence and the sequence generated by some policy, which is an important hyperparameter.%

The L2S approach has a lot of similarities with %
reinforcement learning (RL), and the critical difference is the access to the reference policy that chooses tokens optimally w.r.t.\ the annotation and cost function (such policies relate to the expert oracles of the RL world, where they are usually not available). 
Reference policies are often difficult to construct for non-trivial test metrics like the BLEU  \citep{papineni2002bleu} score, so one resorts to using approximations.

In this paper, we investigate different components of the learning-to-search approach (reference policy, costs and loss function) on the tasks of word ordering, neural machine translation (NMT), code generation.  
First, we propose the reference policy based on the alignment between the predictions and ground truth and prove its optimality w.r.t.\ the Kendall-tau distance between the permutations (appear in the word ordering task).
The constructed alignment can also be used to approximate the METEOR score \citep{denkowski2014meteor}. 
Next, we experiment with different costs for word ordering, code generation and neural machine translation and show that the learning-to-search approach benefits from  choosing the costs related to the test metrics. Although it is expected behavior, this is not always the case \citep{Shen2016b, wieting2019beyondbleu}. 

For the training objective, \citet{searnn2018leblond} use the KL divergence between the distributions obtained from the costs and the model. %
With this loss, we observe that  training  requires an extreme value of the parameter controlling the scale of costs, which corresponds to the degenerate target distribution. In this mode, the model learns only few (1-2) tokens corresponding to the lowest costs  and expensive computation of other costs appears to be useless. 
We consider alternative training losses that depend only on the ordering of costs. First, we redefine the target distribution  for the  KL loss based on the cost ordering. Second, we adapt the top-$k$ ListMLE loss from the ranking literature~\citep{xia2009statistical} to the L2S pipeline. %
We report that the ordering-based losses can outperform the original KL loss.

The  rest of the paper is organized as follow: we describe the training with MLE and SeaRNN objectives in the Section \ref{training_L2S} and we give the details of the experiments in the Section \ref{setup}. In the Section \ref{cost_sensitive_section}, we choose the test metrics and define the costs and reference policy based on these metrics; in the Section \ref{losses_section}, we discuss the training with different losses. The Sections \ref{related_work} and  \ref{conclusion} provide the related work and conclusion respectively.

\section{Training with learning-to-search}
\label{training_L2S}
 Consider an  \textit{autoregressive} model with the input $x$ and sequential output $(y^{1},\ y^{2},\ \dots \ ,\ y^{\tau})$. At each step, the goal is to predict a new output $y^{t+1}$ given the previous outputs $(y^{1},\ y^{2},\ \dots \ ,\ y^{t})$ and the input $x$. Such models are usually trained with \textbf{maximum likelihood estimation}, which consists in maximizing the following objective:
\begin{equation}
\label{MLE_loss}
  \log p(\mathbf{y} | \mathbf{x},\theta) = \sum\limits_{t = 0}^{\tau - 1} \log p(y^{t + 1} | y^{1}, \dots ,y^{t},\mathbf{x},\theta ). %
\end{equation}
A common practice is to model the conditional probability $p(\mathbf{y} | \mathbf{x}, \theta)$  by a neural network with an encoder-decoder architecture. The encoder maps the input to the latent space, which the decoder uses to produce the output. The encoder and decoder are usually build with the RNN~\citep{Bahdanau2015} or Transformer~\citep{Vaswani2017} blocks. 

At the training stage, the input $y^{t-1}$ of the model  is the ground-truth token from the previous step (\textit{teacher forcing}). At the testing stage, the ground-truth tokens are not available, so the input of the model is the output from the previous step. %

\subsection{SeaRNN algorithm}
One alternative to the MLE training of  autoregressive models is  the learning-to-search approach \citep{Daume2009b, Chang2015, searnn2018leblond}. We use the \textbf{SeaRNN} algorithm  \citep{searnn2018leblond}.  For computing the loss for the $t$-th token ($t=1, \dots, T$) of an autoregressive model, the algorithm does the following steps:     
    \begin{enumerate}        
        \item Construct the prefix of  length $t$  according to the  \textit{roll-in policy} of choosing  tokens.
        \item Sample $k$ tokens $y_1, \dots, y_k$ to try. For the tasks with small vocabularies,  we can sample all available tokens, which is too expensive otherwise. Sampling can be done   uniformly from all available tokens, by choosing some fixed number of neighbors of this token in the ground truth or by selecting the tokens with the top-$k$ values of probability according to the current model.
        \item  Try adding each token to the prefix and complete all the sequences with the \textit{roll-out policy} of choosing  tokens. 
        \item Measure how close each completed sequence is to the ground truth, obtain the  \textit{costs} and compute the training loss. 
    \end{enumerate}
    
A popular choice of such training loss is the KL-divergence between the distributions defined from the costs and model (Section \ref{origin_kl}).
The costs are typically defined according to the test metric, inform the model how good or bad were the choices of tokens and allow the SeaRNN algorithm  to optimize the test metric during the training stage.

\textbf{Roll-in and roll-out policies.}
An important component of the method is the policy of choosing the tokens. We need these tokens to construct a prefix (roll-in) and to complete a sequence (roll-out); the completed sequences are used for computing the costs. %
Differently to the MLE training with teacher forcing, when using SeaRNN, the selected tokens are used as the input of the model.

Both roll-in and roll-out policies can be \textit{reference}, \textit{learned} or \textit{mixed}.
The \textit{learned} policy chooses the most probable word according to the current model. 
The \textit{reference} policy at roll-in acts as teacher forcing and always outputs the ground-truth tokens. At roll-out, the reference policy  completes the roll-in prefix optimally w.r.t.\ the test metric. 
The optimal reference policy w.r.t.\  metrics like BLEU or METEOR is hard to define, so we need to approximate it (we also refer to these approximations  as the reference policy). 
Finally, we can mix the reference and learned policies. We choose reference or learned policy with probability $p$ for each sequence (\textit{mixed} policy) or for each step (\textit{mixed-cells} policy). 

\section{Experimental setup}
\label{setup}
We now describe the tasks, datasets and models used in our experiments (Sections \ref{cost_sensitive_section} and \ref{losses_section}).
\subsection{Tasks and Datasets}
\textbf{Word ordering.}
The goal of this task is to recover the order of words from the permuted sentence. We use the English part of the Multi30k dataset \citep{Elliott2016}. Following \citep{Gu2019InsertionbasedDW}, we randomly permute the sentences to obtain the inputs.

\textbf{Code generation.} The task is to generate Python code from the descriptions in natural language. We use the standard sequence-to-sequence framework and the Django dataset \citep{Oda2015} .

\textbf{Neural machine translation.} The task is to translate sentences from a source language to a target language. The neural machine translation experiments are conducted on the Multi30k dataset \citep{Elliott2016}; the source language is German and the target one is English.

\subsection{Models}
In all experiments, we use the standard encoder-decoder architecture with the attention of \citet{Bahdanau2015}, gated recurrent units \citep{Cho2014} and bidirectional encoder. The models have 2 layers and are regularized with the dropout of rate 0.3. We train the models with the Adam optimizer with the learning rate of $10^{-3}$. For inference, we use the greedy decoding. 
For the word ordering task, we constrained the model to always output a permutation of the input (by masking the output of the softmax layer). Other hyperparameters and training details are provided in Appendix \ref{hyperparameters}.

\section{Cost-sensitive training}
\label{cost_sensitive_section}
In this section, we describe the test metrics that we use, and the way to define the costs and reference policy based on these metrics. 

We start with the BLEU metric, the corresponding costs and reference policy (Section \ref{training_bleu}). Next, we describe the Kendall-tau distance and propose the reference policy, which we prove to be optimal w.r.t.\ this metric (Section \ref{training_kendall}). In the Section \ref{training_meteor}, we describe the METEOR metric and propose the reference policy and the way to approximate the METEOR score for computing the costs. Finally, we train the models to optimize these metrics and report the results (Section \ref{costs_results}).

\subsection{Training with BLEU} 
\label{training_bleu}
BLEU \citep{papineni2002bleu}  is a widely-used metric for text generation tasks. This metric is based on $n$-grams of different $n$ (the standard choice is BLEU-4 with $n \le 4$). All our tasks is measured with BLEU: word ordering \citep{Wiseman2016b},  machine translation \citep{papineni2002bleu}, code generation \citep{Oda2015}. When training to maximize BLEU, we use the reference policy proposed by \citet{searnn2018leblond}: they try adding every suffix in the ground-truth sequence to the current prediction and pick the one with the highest BLEU-1 score.
They follow \citet{Bahdanau2016} and use a sentence-level smoothed version of BLEU-4 as the costs.

However, the  BLEU is known to have drawbacks: it does not rely on the word meaning or  grammatical structure \citep{callison-burch-etal-2006-evaluating}; it correlates with the human evaluation less than other metrics \citep{sun-2010-mining}; it is difficult to optimize \citep{wieting2019beyondbleu}. We consider alternative metrics for all our tasks and investigate how different cost functions influence the results of training.

\subsection{Training with Kendall-tau} 
\label{training_kendall}
For the word ordering task, we use the Kendall-tau distance as the additional to BLEU metric. 
The Kendall-tau distance is a standard way to measure the difference between two permutations: it computes the number of pairwise disagreements between the two permutations. 
The only difference between the predicted  and ground-truth sequences in  the word ordering task is the order of tokens, which means that we can consider the predicted sequence as a permutation of the ground truth. 
Specifically, we use the Kendall-tau distance between the permutation, which corresponds to the predicted sequence, and the identity permutation, which corresponds to the ground-truth sequence. 

The reference policy of SeaRNN can not be used for the word ordering task because it does not return a permutation of the input. We propose the reference policy based  on the alignment between the ground-truth sequence and the current predictions. 
The proposed policy completes the current predictions with missing elements and adds them in the order they appear in the ground truth. 
We prove that this reference policy is optimal w.r.t.\ the Kendall-tau distance (Appendix \ref{appendix_opt_policy}).

\subsection{Training with METEOR} 
\label{training_meteor}
For the neural machine translation task, METEOR \citep{denkowski2014meteor}, as well as BLEU, is a popular choice for the quality estimation. METEOR has an alignment module for computing the score. 
The alignment, constructed by METEOR, maximizes the number of covered tokens and minimizes the number of chunks (here a chunk is a contiguous subsequence with the correct internal order). 

While the direct computation of METEOR at each training step is computationally expensive, we can construct the alignment similarly to Section \ref{training_kendall} and define the simplified version of this metric as a cost function (we call it sMETEOR). 
We extend the proposed alignment reference policy: instead of adding the missing elements in the order they appear in the ground truth, we group them into chunks (see Appendix \ref{meteor_alignment} for the details). 
We add missing chunks in the order that minimizes the number of chunks, as in the METEOR computation. 

The standard METEOR score still computes the alignments in a more sophisticated way: it allows to align not only the tokens matching exactly, but also with other types of matching. 
For simplicity, we allow only exact matching in sMETEOR. For the code generation task, we use only the exact matching in the standard METEOR metric (we refer it as METEOR-e.m.).

\subsection{Results}
\label{costs_results}
We apply the proposed in the Section \ref{training_kendall} alignment policy to the word ordering task and compare the two settings: training with the BLEU  and Kendall-tau costs. The results show that learning-to-search outperforms MLE for both metrics and the training method benefits from using the correct cost function (Table \ref{wo_costs}).%

In the neural machine translation and code generation tasks, our experiments show that even with our simplified version of METEOR, sMETEOR (defined in Section \ref{training_meteor}), as costs and our alignment policy, we can maximize the original METEOR metric (Table~\ref{nmt_cg_costs}). The results with learning-to-search outperform MLE. However, difference between the training with BLEU and sMETEOR costs is not very large in terms of the METEOR metric.  This is probably due to the fact  that code generation and neural machine translation are more challenging tasks than word ordering because there are a lot of factors that affect training. %

\begin{table}[t]
\caption{Results  on the word ordering task with the costs based on  BLEU and the Kendall-tau distance. For each method, we report both BLEU and the Kendall-tau distance on the test set.}
\centering
\begin{tabular}{@{}l@{$\:$}r@{$\;\;$}r@{}}
&&\\
\toprule
   Training Method & BLEU $\uparrow$ & Kendall-tau $\downarrow$    \\ \midrule
MLE          & 53.98     & 15.20         \\
BLEU costs  & \textbf{55.19} & 15.19    \\ 
Kendall-tau costs & 50.74          & \textbf{14.30} \\
\bottomrule
\end{tabular}
\label{wo_costs}
\end{table} 

\begin{table}[t]
\caption{Results on the neural machine translation and code generation tasks with the costs based on BLEU and METEOR.  For each method, we report both BLEU and METEOR on the test set.}
\centering
\begin{tabular}{@{}lcccc@{}}
              & \multicolumn{2}{c}{NMT} & \multicolumn{2}{c}{Code generation} \\ \midrule
Training Method         & BLEU $\uparrow$       & METEOR   $\uparrow$  & BLEU $\uparrow$            & METEOR-e.m. $\uparrow$         \\ \midrule
 MLE              &  38.71      &    36.73         & 51.46           & 70.42           \\
BLEU costs   & \textbf{40.52}    & 37.76  & \textbf{60.69}           & 75.84           \\
sMETEOR costs     & 39.86     & \textbf{37.87}    & 59.60           & \textbf{76.29}          
\end{tabular}
\label{nmt_cg_costs}
\end{table}

\section{The ranks of the costs are more important than the values}
\label{losses_section}

In the previous section, we discussed  training with different costs, which could be of different scales. In the original KL loss, the scale of the costs is controlled with the parameter, which is important and requires tuning (Section \ref{origin_kl}). To evaluate the importance of the scale of costs, we try two ordering-based losses. The first one replaces the distribution formed from the cost values with the distribution formed from the cost order (Section \ref{ordinary_kl}). The second one is based on a loss from learning-to-rank (Section \ref{ranking_loss}). We compare training with different losses and investigate what is more important for training: ordering or scale. We discuss the effect of the scale parameter on the quality of the model in Section \ref{top_learning}.

For the next experiments, we choose the type of costs that provide larger performance improvement w.r.t.\ MLE for each task: the Kendall-tau costs on the word ordering task, and  BLEU on the neural machine translation and code generation tasks.

\begin{table}[t]
\caption{Results of the training on word ordering, neural machine translation and code generation with different losses. We report the Kendall-tau distance for word ordering, BLEU for code generation and machine translation (all on the test sets).
}
\centering
\begin{tabular}{lrrrrrrrr}
\toprule
            & MLE   & KL    & KL (q 0.9)  & KL (q 0.7) & top-1 ListMLE & top-2 ListMLE   \\ \midrule
Word ord. $\downarrow$ &15.20    &14.30 & \textbf{14.03}    &14.05 & 14.15   & 14.14 \\
Code gen. $\uparrow$& 51.46  & \textbf{60.69}  &    58.90  & 59.40   & 59.42 &  40.41 \\
NMT $\uparrow$ & 38.71  &  40.52 & \textbf{40.72}    & 40.53    &   40.28  & 38.94  \\

\bottomrule
\end{tabular}
\label{losses_table}
\end{table}
\subsection{Original KL loss} 
\label{origin_kl}
The widely-used objective that includes the costs is the KL-divergence between the model and cost distributions  at each prediction step \citep{searnn2018leblond,  NIPS2018_7820, sabour2018optimal, welleck2019non}.
The model distribution is the model output, which corresponds to the probability of each token in the dictionary at the current step. The standard way to convert the costs into a distribution is to use the softmax function. 
The costs can be of different scales in different cases, which affects the target distribution. An additional scale parameter (the inverse of the temperature) controls the scale of the costs.

The KL loss up to a constant equals the cross-entropy between the model and cost distributions  at each prediction step: %
\begin{align}
\label{original_KL_loss}
	\mathcal{L}_t(\text{costs}_{t}, \text{model}_{t}) &= -%
	\sum\limits_{i =1}^k p^{\text{costs}}_{t}(a) \log\big(p_t^{\text{model}}(a)\big), \\
\label{p_costs}
 p^{\text{costs}}_t(a) &= \dfrac{\exp({-\alpha\cdot \text{cost}_t(a)}) }{\sum\nolimits_{i =1}^k \exp({-\alpha \cdot \text{cost}_t(i)}) }, \\
 p_t^{\text{model}}(a) &=\dfrac{ \exp({\text{scores}_t(a)})}{\sum\nolimits_{i =1}^k \exp({\text{scores}_t(i)})}.
 \end{align}
Here $\alpha$ is the scale parameter, $k$ is the number of samples. 
The loss \eqref{original_KL_loss} relies on the scale of the cost function and it is important to tune the scale parameter, which appears to be highly correlated with the model quality (Section \ref{top_learning}).

\subsection{Ordering-based KL loss}
\label{ordinary_kl}
Instead of using the target distribution obtained with the softmax function of the costs, we define the target distribution from only the ordering of the costs. We parameterize the target distribution with one parameter $q$ similarly to the stick-breaking process:
\begin{align*}
\label{ordinary_KL_loss}
    p^{\text{costs}}_t(a_{\pi_i}) &= q (1 - q)^{i- 1},\ i \in [1 \mathinner{\ldotp \ldotp} k-1], \\
    p^{\text{costs}}_t(a_{\pi_{k}}) &= (1 - q)^{k-1}.
\end{align*}
Here $q \in (0.5 \mathinner{\ldotp \ldotp} 1]$ and $\pi$ is a permutation corresponding to the non-decreasing order of the cost values ($\pi_1$ gives the index of a token with the smallest cost value). 

\subsection{Top-$k$ ListMLE loss}
\label{ranking_loss}
We can explicitly learn the order of the cost values with the top-$k$ ListMLE loss \citep{xia2009statistical}, which comes from learning-to-rank. The order of costs corresponds to the ground-truth permutation that we want to learn.

The top-$k$ ListMLE loss \citep{xia2009statistical} is the negative log-likelihood of the  top-k subgroup in the ground-truth permutation :
\begin{equation}
    \mathcal{L}_t(\text{costs}_{t}, \text{model}_{t}) = - \log \prod\limits_{i=1}^{top\text{-}k} \dfrac{\exp(a_{\pi_i})}{ \sum_{j=i}^k \exp(a_{\pi_j}) }.
\end{equation}
When $k$ equals one, this loss is equivalent to  target learning used by \citet{searnn2018leblond}.

\begin{figure}
\begin{subfigure}{.5\textwidth}
  \centering
  \includegraphics[scale=0.28]{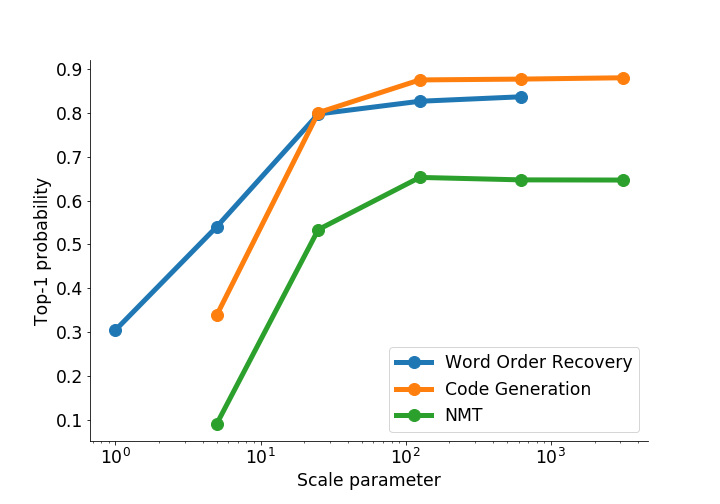}
  \caption{The top-1 probability of the model distribution}
\end{subfigure}
\hfill
\begin{subfigure}{.5\textwidth}
  \centering
  \includegraphics[scale=0.28]{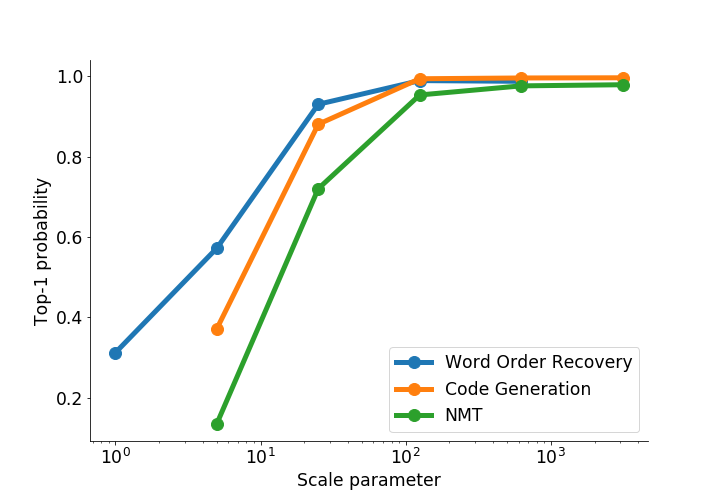}
  \caption{The top-1 probability of the cost distribution}
\end{subfigure}
\caption{Dependence between the scale parameter and the probability of top-1 token according to costs and model.}
\label{pic_top-1}
\end{figure}

\subsection{Results and discussion}
\label{top_learning}
We compare all the losses of Sections \ref{origin_kl}, \ref{ordinary_kl}, \ref{ranking_loss} and observe that for the word ordering and machine translation tasks the ordering-based losses  perform better than the original KL and MLE losses (Table \ref{losses_table}): best results are achieved with ordering-based KL loss ($q = 0.9$).
Information about the ordering of the costs appears to be sufficient for training.
Let us discuss why the performance does not improve when using all the values of the costs in the original KL loss \eqref{original_KL_loss}. We investigate the importance of the scale parameter in the original KL loss. When increasing this parameter, $p^{\text{costs}}_t$ from \eqref{p_costs} gets close to the degenerate distribution, where all the probability mass is concentrated at one point (Figure~\ref{pic_top-1}). 

We observe that the training with original KL loss \eqref{original_KL_loss} works only for the high values of the scale parameter (the values that give the best performance are in the range of $10^2\text{-}10^3$;  training with the values $>10^{3}$  does not provide improvements; the model trained with the parameter in the range $10^{0}\text{-}10^1$ performs worse).  This means that \textbf{the KL loss learns only  1-2 entries of the target distribution with the lowest costs}. The influence of other costs is very low. 
Computing such costs appears to be effectively useless, while it is the most computationally expensive part of the training process.

\section{Related work}
\label{related_work}
The learning-to-search approach is used in different tasks. \citet{NIPS2018_7820} applied L2S for multiset prediction. Unlike our applications, in multiset prediction, the costs are naturally designed to be equal for all tokens from the multiset and the sizes of the multisets are smaller than the dictionary sizes in our tasks.
\citet{searnn2018leblond} proposed the SeaRNN algorithm and demonstrated its advantages on OCR, spelling correction and neural machine translation.
\citet{sabour2018optimal} used  the L2S approach for speech recognition.
\citet{welleck2019non} used L2S to generate text without pre-specifying a generation order. They also studied several variants of the reference policy for their task. However, these policies do not depend on the test metric directly. 

Some works investigate the components of the alternative to MLE training algorithms. 
While proposing minimum risk training (MRT), \citet{shen-etal-2016-minimum} studied the effect of different costs for neural machine translation. 
They concluded that training with sentence-level BLEU improves the results in terms of the corpus level metrics. They also noticed that training with the costs based on some other metrics does not lead to improvement in the corresponding corpus level metrics.
\citet{wieting2019beyondbleu} claimed that BLEU is hard to optimize and proposed an alternative metric for MRT called SIMILE. They used the costs based on this metric to optimize BLEU to improve the translation quality in terms of both BLEU and SIMILE metrics. 

\citet{reinforce_top19} found that neural machine translation models trained with common RL methods improve the translation quality only when the correct translations are already in the top of the distribution of a pre-trained model used for initialization. They noticed that such an improvement might be easier to achieve with reranking methods instead of RL. Their finding is very close to the effect we see in the model trained with the original KL loss, although the SeaRNN algorithm does not require pre-training. \citet{edunov-etal-2018-classical} investigated different losses from structured prediction in the context of neural machine translation. They found that combining sequence-level and token-level losses performed better. The losses in our work are computed at the token level, but use the sentence level while obtaining the costs. 

\section{Conclusion}
\label{conclusion}

In this study, we investigate the effect of different components of the SeaRNN algorithm \citep{searnn2018leblond}: the reference policy, costs, and loss functions. 
For word ordering, we show that the performance improves when choosing the costs related to the test metrics and training with the optimal to the Kendall-tau distance reference policy, which we proposed.
In the case of optimizing the BLEU and METEOR metrics, the method benefits less from using the correct costs and more work is required to understand this effect. We observe  that  the original KL loss tends to learn only the top tokens of the target distribution and does not fully utilize the costs. We propose the losses based only on  the ordering of the costs and  demonstrate that these losses can perform better than the original KL loss.

\section*{Acknowledgments}
This research is in part based on the work supported by Samsung Research, Samsung Electronics, and by the Russian Science Foundation grant no. 19-71-30020.
We also thank NRU HSE for providing computational resources.

\bibliography{references}
\bibliographystyle{icml2019}

\newpage
\begin{appendices}

\section{Training details}
\label{hyperparameters}

\begin{table}[h]
\caption{Training hyperparameters}
\centering
\begin{tabular}{lrrr}
\toprule
Parameters               & Word ordering  & Code generation           & NMT     
               
               \\
\midrule
max length     & 80         & 50                  & 50                  \\
roll-in        &  mixed          &      mixed-cells  &  mixed-cells                   \\
roll-out       &  mixed          &        mixed          &   mixed                  \\
max iteration       & 25000      & 10000               & 10000               \\
batch size     & 32         & 32                  & 128                 \\
embedding size     & 500        & 128                 & 500                 \\
hidden size    & 500        & 256                 & 500                
\end{tabular}
\label{params}
\end{table}
Additional hyperparameters can be found in Table \ref{params}; we chose these values based on the performance on the validation set. 

When training the word ordering models, we share the source and target embeddings. We choose all tokens to try from ground truth that are not in the prefix. When training the models on the code generation and neural machine translation tasks, we follow \citet{searnn2018leblond} and choose 5 tokens  before and after the current position in the ground truth and 15 tokens that correspond to the top probabilities of the model distribution. The probability of reference or learned policy step is $0.5$ for both mixed and mixed-cells modes \citep{searnn2018leblond}.

We observed that the model receive wrong signal from the costs defined with the default METEOR parameters. Computing the approximate METEOR scores in neural machine translation, we use the following parameters of the metric $\alpha=0.5$, $\beta=2$, $\gamma = 0.5$. For code generation, we use the language-independent METEOR parameters described in the METEOR documentation. 

\section{Optimal policy for the Kendall-tau distance}
\label{appendix_opt_policy}
\paragraph{Proposition.} The reference policy that inserts missing tokens from the ground truth in the order of their positions in the ground truth  is optimal w.r.t.\ the Kendall-tau distance.
\paragraph{Proof.} The ground truth corresponds to the identity permutation $(1, \dots, n)$.  Given the prefix $(\sigma_1, \dots, \sigma_k)$ we seek to obtain the lowest possible value of the Kendall-tau distance between  the ground truth and its permutation $(\sigma_1, \dots, \sigma_n)$. By definition, we can write the Kendall-tau distance as follows:
\begin{equation*}
  \begin{gathered}K = \dfrac{2}{n(n-1)} \sum\limits_{\text{all pairs } (i, j), i \neq j} \mathbb{I} [ (\sigma_i < \sigma_j) \& (i > j) \text{ or } (\sigma_i > \sigma_j) \& (i < j)] = \\
    = \dfrac{2}{n(n-1)} \sum\limits_{\text{all pairs } (i, j), i \neq j} \mathbb{I} (\sigma_i > \sigma_j) = \\
   = \dfrac{2}{n(n-1)} \Bigg[
   \underbrace{\sum\limits_{i \le k, j \le k, i \neq j} \mathbb{I} (\sigma_i > \sigma_j)}_{\substack{\text{the prefix,} \\\text{equal for all completions}}} +
   \underbrace{\sum\limits_{i \le k, j > k} \mathbb{I} (\sigma_i > \sigma_j)}_{\substack{\text{depends only on the set of missing tokens,}\\\text{equal for all completions}}} +
   \underbrace{\sum\limits_{i > k, j >k, i \neq j} \mathbb{I} (\sigma_i > \sigma_j)}_{\substack{\text{suffix,}\\\text{ minimal when no inversions}}} 
   \Bigg].
  \end{gathered}
\end{equation*}
 The last term corresponds to the suffix and is minimal if  the completion has the minimal number of inversions. We can achieve zero inversions, when all tokens in the suffix come in the order they appear in the ground truth. This is exactly what our alignment policy outputs.
  
\section{Alignment policy for the  METEOR metric}
\label{meteor_alignment}
To optimize METEOR, we proposed the policy based on the alignment between the ground-truth sequence and current prediction. 
For this purpose, we construct the alignment at each step of the prediction. %
If  the current policy is reference, it does one of the following steps:
        \begin{enumerate}
            \item if it is the first step, the reference policy returns the first token of the ground truth;
            \item if it is not the first step
            and we can continue the already started chunk, the reference policy returns the next token, which continues this chunk;
            \item if it is not the first step, but we can not continue the started chunk (the next token is already used or the current token is the end-of-sequence), we start a new chunk with the token that is not used yet and appears in the ground truth first.
        \end{enumerate}
If the policy of the current step is learned and the output of the model aligns with the ground-truth token that is not used yet we update our alignment.

\end{appendices}

\end{document}